%% file: report.tex
\documentclass[a4paper, 11pt]{article}

\usepackage[english]{babel}
\usepackage[utf8x]{inputenc}
\usepackage[T1]{fontenc}
\usepackage{times}
\usepackage{latexsym}
\usepackage{microtype}
\usepackage{graphicx}
\usepackage[colorinlistoftodos]{todonotes}
\usepackage[margin=0.8in, top=1in]{geometry}
\usepackage{wrapfig,lipsum,booktabs}
\usepackage{subfig}
\usepackage{multirow}
\title{\vspace{-3em}Ensemble Language Models for Multilingual Sentiment Analysis}
\author{
Md. Arid Hasan\\
\texttt{arid.hasan@unb.ca} \\
\texttt{University of New Brunswick} \\
}
\date{}

\begin{document}
\maketitle

\vspace{-4em}
\section*{ABSTRACT}
\vspace{-1em}
The rapid advancement of social media enables us to analyze user opinions. In recent times, sentiment analysis has shown a prominent research gap in understanding human sentiment based on the content shared on social media. Although sentiment analysis for commonly spoken languages has advanced significantly, low-resource languages like Arabic continue to get little research due to resource limitations.
In this study, we explore sentiment analysis on tweet texts from SemEval-17 and the Arabic Sentiment Tweet dataset. Moreover, We investigated four pretrained language models and proposed two ensemble language models. Our findings include monolingual models exhibiting superior performance and ensemble models outperforming the baseline while the majority voting ensemble outperforms the English language.

\section*{KEYWORDS}
\vspace{-1em}
Sentiment Analysis; Ensemble Language Models; Multilingual BERT; XLM-RoBERTa; RoBERTa

\section{INTRODUCTION}
\vspace{-1em}
In recent years, Natural Language Processing (NLP) has seen remarkable advancements, with sentiment analysis emerging as a pivotal subfield. This analytical tool finds applications across diverse domains such as market research, customer feedback analysis, social media monitoring, and brand reputation management. The ability to decipher sentiments expressed in textual data has proven indispensable for businesses and researchers seeking actionable insights from the vast sea of user-generated content.
This project delves into sentiment analysis with a specific focus on the intricate landscape of multilingual tweet texts. The objective is to discern and classify sentiments within these succinct and dynamic messages into three overarching categories: positive, negative, or neutral. The importance of this task extends beyond linguistic diversity, encompassing a range of domains, including business, finance, politics, education, and various services \cite{cui2023survey}. 
To undertake this endeavor, we leverage existing datasets from SemEval-2017 Task 4, particularly Task 4A, which entails message polarity classification. The richness of this dataset, combining English data from shared tasks spanning SemEval 2013 to 2017, presents a valuable foundation for our sentiment analysis endeavors. However, challenges arise due to linguistic imbalances, particularly in the Arabic dataset, prompting the curation of the Arabic Sentiment Tweet Dataset (ASTD) \cite{nabil2015astd}. By merging ASTD with SemEval-2017 Arabic sentiment data, we aim to mitigate biases and enhance the representation of sentiments in the Arabic language.

Our proposed approach comprises two key facets. Firstly, we plan to harness the power of pretrained language models, including AraBERTv2, RoBERTa, multilingual BERT, and XLM-RoBERTa. These models will undergo individual fine-tuning processes with both English and Arabic datasets, optimizing them for the nuanced linguistic characteristics of each language. In the second facet, we propose an ensemble approach. After individual fine-tuning, these models will be amalgamated into a language-independent ensemble model. This model, trained on the combined dataset, is expected to provide a holistic and robust sentiment analysis solution that transcends linguistic boundaries.
To benchmark the effectiveness of our approach, we plan a comprehensive evaluation against State-of-the-Art Deep Learning Models. This comparative analysis will utilize the macro-average F1 measure as the evaluation metric, chosen for its suitability in assessing performance in multi-class imbalanced datasets \cite{antoun2020arabert, nabil2015astd}.

In summary, this project stands at the intersection of NLP, sentiment analysis, and multilingual understanding, aiming to contribute insights and methodologies that advance the field and offer practical solutions for real-world applications.

\section{BACKGROUND}
\vspace{-1em}
In the contemporary era of prolific social media usage, sentiment analysis has become integral, especially on microblogging platforms like Twitter. The wealth of user-generated data presents a fertile ground for understanding public opinions, feedback, and sentiments \cite{6897213}.
One prevalent approach in sentiment analysis involves leveraging machine learning algorithms for classification, as demonstrated in studies such as the one utilizing distant supervision and noisy labels from emoticons and acronyms in Twitter messages \cite{6897213}. This methodology addresses the brevity and simplicity inherent in tweets, contributing to the nuanced analysis of sentiments.

Linguistic diversity, cultural nuances, and geographically specific trends on social media platforms add complexity to sentiment analysis. Cross-lingual and multilingual approaches have emerged to tackle this challenge, exemplified by the research utilizing XLM-RoBERTa for cross-lingual sentiment analysis, transferring knowledge from resource-rich English to resource-poor Hindi \cite{10.1145/3461764}. Such adaptability is crucial for effective sentiment analysis across diverse linguistic landscapes. Furthermore, sentiment analysis is not limited to English-centric platforms. Research on sentiment detection in Arabic tweets showcases efforts in combining pre-processing strategies with transformer-based models like AraELECTRA and AraBERT, addressing the nuances of sarcasm and sentiments in the Arabic language \cite{wadhawan2021arabert}.

The surge in social media usage, particularly on platforms like Twitter, has led to a tremendous increase in user-generated data. This growth necessitates effective text categorization and sentiment analysis, crucial for various domains such as healthcare and policy-making. The multilingual nature of social media data poses a challenge, prompting the exploration of domain-agnostic and multilingual solutions \cite{manias2023multilingual}. Recent endeavors in multilingual text categorization and sentiment analysis involve the utilization of BERT-based classifiers and zero-shot classification approaches. These approaches demonstrate promising accuracy and efficiency in classifying sentiments across diverse languages. Comparative analyses reveal the strengths of multilingual BERT-based classifiers and the versatility of zero-shot approaches in creating efficient and scalable multilingual solutions. The robustness of language models, particularly transformer-based architectures like RoBERTa, has been a focus of research. 
Studies delve into aspects like pretraining techniques, hyperparameter choices, and training data size to optimize performance \cite{liu2019roberta}. The application of RoBERTa in aspect-category sentiment analysis illustrates its superiority over LSTM-based methods, showcasing its potential in extracting nuanced sentiments \cite{liao2021improved}.

This comprehensive background sets the stage for our project, emphasizing the need to explore and compare diverse sentiment analysis methodologies, leverage pretrained language models, and address linguistic challenges across different languages and domains. The proposed research draws inspiration from the insights provided by existing studies and seeks to enhance the understanding and applicability of sentiment analysis techniques in real-world scenarios.

\section{METHODOLOGY}
\vspace{-1em}
In this study, we focus on sentiment analysis of tweet data. Sentiment analysis is used to identify the polarity of a given text. It is usually used to identify the quality of products, customer service, and social media text analysis.

\subsection{Data Collection and Preprocessing}
We mainly used datasets from two sources. We used the dataset from SemEval-17 \cite{rosenthal2019semeval} for English and Arabic data was curated from SemEval-17\cite{rosenthal2019semeval} and ASTD \cite{nabil2015astd}. The datasets were developed using X (formerly named Twitter) data. The ASTD dataset has 4 classes (Objective, Positive, Negative, and Neutral) and the SemEval-17 dataset has 3 classes (Positive, Negative, and Neutral). We remove one class (Objective) from the ASTD data for consistency. For English training data, we combined all the previous data from SemEval 2013-2016 and excluded the test data from 2013 and 2014. We considered the test data of SemEval 2013 and 2014 as the development set for our study and SemEval-17 test data was used for evaluating the performances of the model. For the Arabic data, we used the data from subtasks A, B, and D from SemEval-17 and merged it with the ASTD dataset. From the merged data, we used 90\% data as training and 10\% data for validation of the system. We used the official test set of SemEval-17 subtask A as our test set.

Tweet data always contains many symbols, URLs, usernames, and invisible characters. We preprocess the data by removing symbols, URLs, and invisible characters. As a part of preprocessing, we applied Byte-pair encoding (BPE) tokenizers which were included with pretrained transformer models to tokenize the text.

\subsection{Model Selection and Baseline}
We used 4 different transformer-based pretrained language models that are ArabicBERT version 02 \cite{antoun2020arabert}, RoBERTa base \cite{liu2019roberta}, multilingual BERT \cite{devlin2018bert}, and XLM-RoBERTa base \cite{conneau2019unsupervised}. 

We selected the multilingual BERT model as a baseline system for our study. The reason behind choosing multilingual BERT is a smaller and simpler model compared with the selected four models. Another reason multilingual BERT supports both Arabic and English languages where ArabicBERT is specially designed for the Arabic language and RoBERTa is for the English language. We didn't choose XLM-RoBERTa due to its complexity and more training parameters compared with multilingual BERT.

\begin{figure}[!ht]%
    \centering
    \subfloat[\centering Ensemble of two pretrained language models followed by a multi-head attention and a Feed-Forward Network]{{\includegraphics[width=0.45\textwidth]{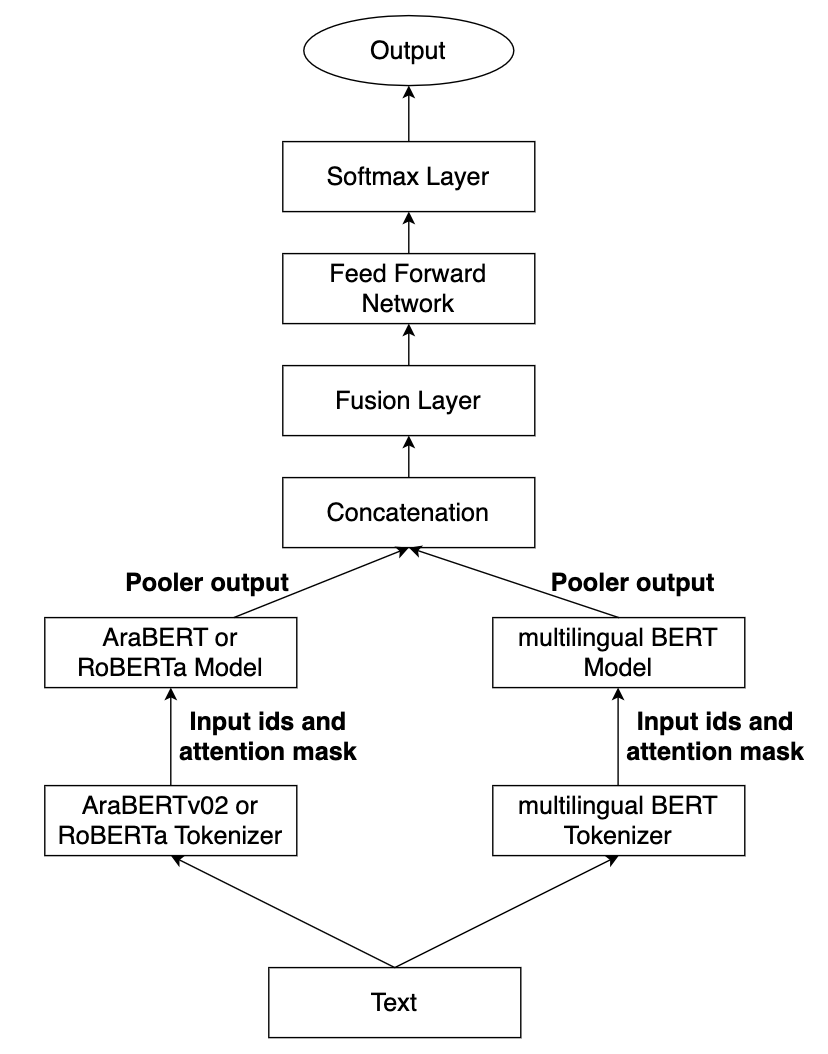} }}%
    \qquad
    \subfloat[\centering Ensemble of two pretrained language models followed by a Feed-Forward Network]{{\includegraphics[width=0.4\textwidth]{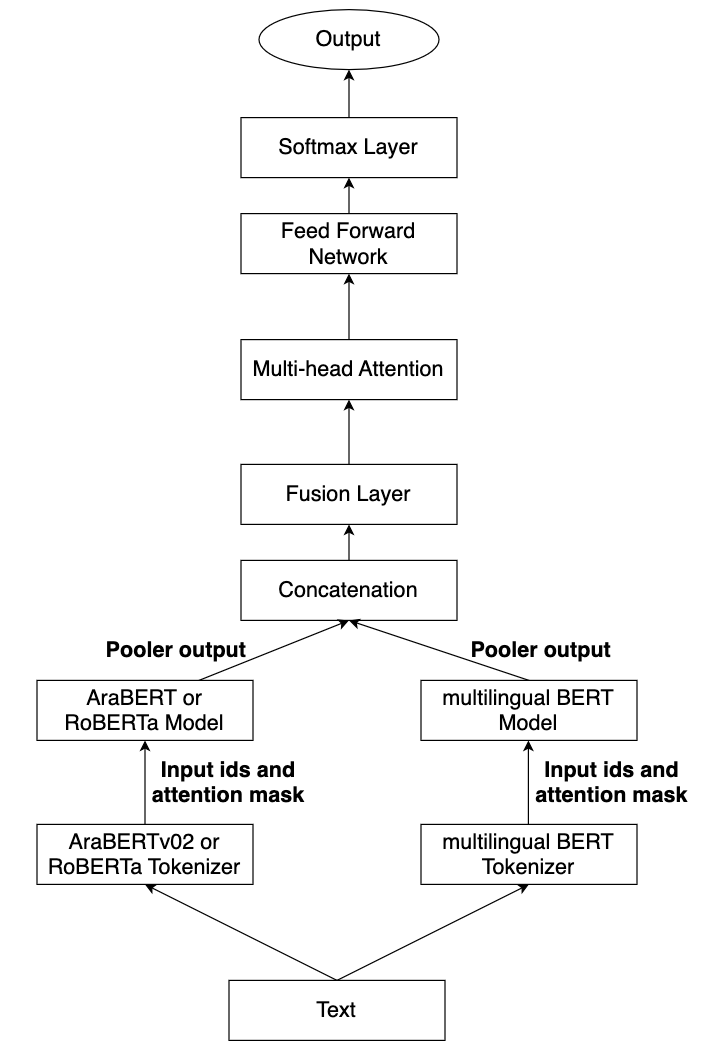} }}%
    \caption{Proposed ensemble models}%
    \label{fig:proposed_models}%
\end{figure}

Although we selected four pretrained models for our study, we also propose two ensemble models to conduct this study. For the first proposed model, we used model model-specific tokenizer as well as a language-specific model. We used the AraBERT model for the Arabic language and the RoBERTa model is used for the English language. We feed the input IDs and attention mask from the tokenizer to the model. Then we extract the pooler output from the language-specific model and multilingual BERT model for concatenation. We pass the concatenated pooler output to the fusion layer followed by the feed-forward network and softmax function. We select the output from the softmax function. For the second proposed model, we applied a multi-head attention layer in between the fusion layer and the feed-forward network. The detailed architecture of the model is presented in Figure \ref{fig:proposed_models}.

\section{IMPLEMENTATION DETAILS}
\begin{figure}[!ht]%
    \centering
    \includegraphics[width=0.6\textwidth]{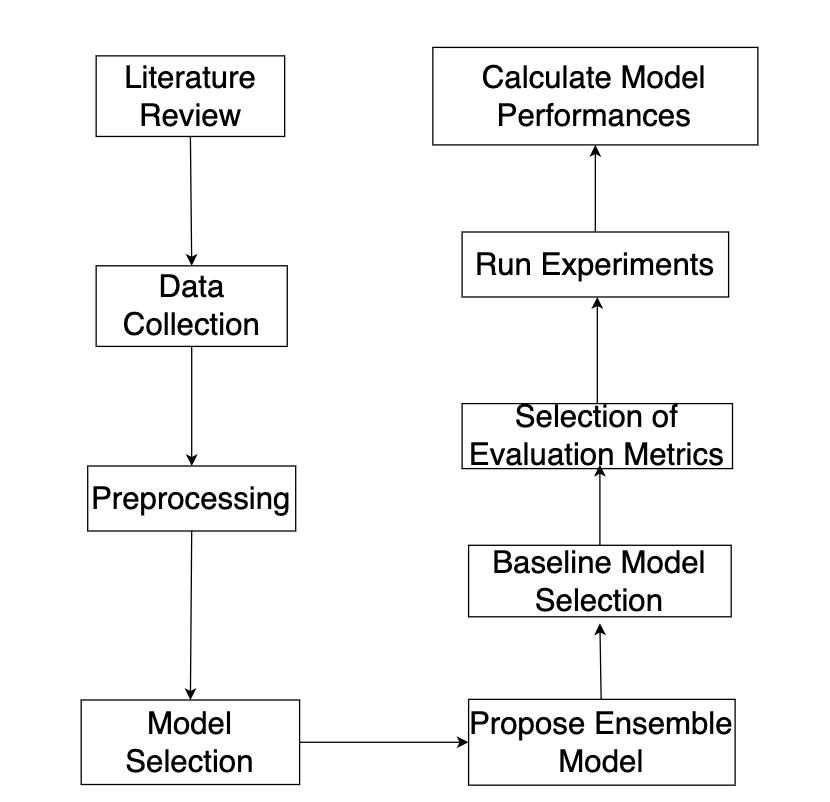}
    \caption{Detailed implementation workflow diagram of our study.}
    \label{fig:workflow}%
\end{figure}

\vspace{-1em}
We first studied two articles to investigate the sentiment analysis problem. We identified the dataset from the articles that we studied. We selected two languages and two different datasets for the Arabic language. We merged the Arabic language datasets into one dataset. Once we completed the data collection and preprocessing, we started to look into the state-of-the-art models. As a result, we selected four pretrained language models to investigate using our dataset. Along with four selected language models, we also propose two ensemble models to investigate. Then we ran several experiments with the selected and proposed models using the datasets. After that, we identified the baseline model for our study. Then, we identified and curated our evaluation metrics for this study. Later, we calculate the performance of each model that we investigate. The detailed workflow is shown in figure \ref{fig:workflow}.

We used a few libraries and two machine learning frameworks to implement our models. First, we selected the PyTorch machine learning framework for writing codes. Although many frameworks are available in the literature, the reason behind choosing the PyTorch framework for its scalability. We can also change the network behavior at runtime while using PyTorch. The second machine learning framework that we use is Transformers by Huggingface. We choose Transformers for its built-in pretrained language models. It's also time and resource consuming to develop a language model from scratch. Moreover, the Transformers framework provides interoperability for PyTorch. We also used the datasets library which enables us to deal with the data easily for fine-tuning language models. The Evaluate library is used to evaluate the fine-tuned models easily. Moreover, we also use scikit-learn to calculate the evaluation metrics described in section \ref{ss:metrics}. We also use the CSV, OS, random, and pandas in our study to read and write the data files. We used the re library for removing the URLs with the help of regular expression.

\section{EVALUATION}

\subsection{Experimental Setup}
We run the experiments into three different settings using the four selected pretrained language models and two proposed ensemble models.
\subsubsection{First Experiment Setup}
In this experiment set, we split the experiments based on languages. We fine-tuned two models for both Arabic and English and used one language-specific model to identify the performances between multilingual and monolingual models. For the English language, we fine-tune the RoBERTa base as the monolingual model, multilingual BERT as the baseline system, and XLM-RoBERTa base model. For the Arabic language, we fine-tune AraBERT as the monolingual model, multilingual BERT as the baseline system, and XLM-RoBERTa base model. We use binary cross entropy with logit loss as the loss function, a learning rate (Adam) of $2e^{-5}$, the maximum sequence length of $256$, batch size of $16$, and then we run $3$ epochs to train all the models.

\subsubsection{Second Experiment Setup}
In this experiment set, we first combined the English and Arabic train sets, validation sets, and test sets to train the model for only one time. We first train the ensemble model with a feed-forward network using the merged trained data. We validate our model for every epoch. Finally, we evaluate our proposed model using the merged test set and calculate the performances individually.
We use the cross-entropy loss as the loss function, a learning rate (Adam) of $2e^{-5}$, the maximum sequence length of $256$, batch size of $24$, and then we run $2$ epochs to train both ensemble models.

\subsubsection{Third Experiment Setup}
In this experiment set, we first train our proposed ensemble model with a feed-forward network using the English training set and validate the model using the validation set. We evaluate the model using the English language test set. Then we train the same ensemble model with a feed-forward network using the Arabic training set and validate the model using the Arabic validation set. Later, we evaluate the model using the Arabic test set. We validate our model for every epoch we run. The reason behind the third experiment setup is to find out any differences in performance in comparison with the second experiment setup.
We use the cross-entropy loss as the loss function, a learning rate (Adam) of $2e^{-5}$, the maximum sequence length of $256$, batch size of $24$, and then we run $2$ epochs to train both ensemble models.

\subsection{Evaluation Metrics}
\label{ss:metrics}
For the performance measure for all different experimental settings, we compute accuracy, weighted precision, weighted recall, and macro F1 score. We choose to use the weighted and macro version of the metric as it takes into account class imbalance.

\subsection{Experimental Results}
\label{ss:exp_results}
In Table \ref{tab:results}, we presented the results of our experiments. From the results, the monolingual AraBERTv02 outperforms the Arabic language and the majority voting ensemble outperforms the English language. Our proposed model outperforms the baseline results for both languages.
\begin{table*}[!ht]
    \centering
    \caption{Performances on different sets of experiments including the baseline. \textbf{Bold} indicates the best system for languages.}
    \input{results}
    \label{tab:results}
\end{table*}

\section{CHALLENGES}
\vspace{-1em}
We face some challenges while studying this problem. Although SemEval-17 data is publicly available, we couldn't download the test set due to the unavailability of resources. To solve this issue, we went through all the participants' papers for SemEval-17 task 4. Finally, we obtained the data from QCRI. The second issue that appeared is running experiments using pretrained language models on the CPU takes more time. Moreover, it was difficult to run the experiments using the proposed ensemble models on 12GB memory. To solve this issue, we used a distributed GPU server. To run ensemble models, we used P100 GPU, and T4 was used to run the other models. Another challenge that we faced during the experiments was batching the data for ensemble models due to considering the language-specific pretrained model. To solve this issue, we separated the English and Arabic data and trained the data sequentially.

\section{INDIVIDUAL CONTRIBUTION}
\vspace{-1em}
In the pursuit of this sentiment analysis project, both Arid and Asfia played integral roles, leveraging their unique strengths and expertise. The collaborative effort was fueled by shared enthusiasm and a unified vision for the project's success.

\subsection{Project Inception}
Arid initiated the project by proposing a compelling idea, sparking collaborative discussions that led to the finalization of the project's overarching task. Together, they deliberated on the scope and intricacies of the sentiment analysis endeavor, aligning their perspectives to shape the project's direction.

\subsection{Proposal}
During the composition of the project proposal, Arid took charge of crafting the Overview, Proposed Approach, and Evaluation Metric sections. This involved distilling the project's essence, outlining the proposed methodologies, and establishing the evaluation framework. Simultaneously, Asfia assumed responsibility for detailing the Dataset, Baselines, and Benchmarks sections, ensuring a comprehensive and well-rounded proposal.

\subsection{Report Writing}
As the project progressed to the report-writing stage, the workload was again distributed to leverage the strengths of each team member. Asfia took charge of composing sections 1, 2, 7, and 8, providing clarity and depth in the "Introduction," "Literature Review," and "Conclusion." On the other hand, Arid skillfully authored sections 3, 4, 6, 5, and 9, contributing to the "Abstract", "Methodology," "Implementation," "Evaluation," and "Challenges" sections as well as proofreading the report. Section \ref{ss:exp_results}, detailing the "Experimental Results" was a collaborative effort, showcasing the seamless teamwork of both members.

\subsection{Implementation}
In the realm of implementation and code writing, the team members again adopted a collaborative approach. Asfia led the development of simple models. Arid, meanwhile, implemented fine-tuning models to run both GPU and CPU devices as well as the ensemble models. Moreover, Arid runs all the experiments and generates results.

\section{CONCLUSION}
\vspace{-1em}
In this comprehensive sentiment analysis project, we navigated the intricate landscape of multilingual tweet texts, employing a diverse set of models, datasets, and ensemble strategies to decipher sentiments effectively. The evaluation results, as presented in the tabulated performance metrics, reflect the nuanced challenges and successes encountered across different linguistic and model scenarios. The individual model assessments revealed notable disparities, with RoBERTa exhibiting robust performance in English sentiment analysis, outshining its counterparts. Conversely, m-BERT demonstrated varied efficacy, emphasizing the sensitivity of model choice to language nuances. AraBERTv02 showcased commendable accuracy in Arabic sentiment analysis, albeit facing challenges posed by the language's unique characteristics.

The power of ensemble methods emerged prominently, with Majority Voting Ensemble and the proposed Ensemble multiple BERT approaches demonstrating improved performance, mitigating individual model limitations. The intricacies of language-independent models were explored through combined English and Arabic datasets, highlighting the potential for broader applicability. Despite variations in model performance, the consistent use of macro-average F1 as an evaluation metric allowed for a balanced assessment across imbalanced multi-class datasets. This choice emphasized the project's commitment to precision, recall, and overall model effectiveness in handling diverse sentiments.

In conclusion, this project delved into the dynamic realm of sentiment analysis, offering insights into the effectiveness of state-of-the-art language models across distinct languages. The exploration of ensemble techniques and language-independent models underscored the adaptability and potential generalization of sentiment analysis systems. As sentiment analysis continues to evolve, this project contributes valuable perspectives, paving the way for further advancements in understanding and interpreting sentiments across multilingual contexts.

\section{ETHICS STATEMENT}
\vspace{-1em}
We sampled a few pieces of data to check the annotation quality and found some of the data were not annotated correctly. Due to the biases of human annotators, some of the annotations were biased for the different perceptions on the specific topic. Moreover, the source of the data didn't disclose the annotation agreement score to ensure the quality of the annotation. Although our model provides a good performance on test data, the wide use of the model has yet not been explored. There will be always some scenarios where our model can not predict the actual sentiment class. The performance of the model should be taken into account for public use. We could not analyze the errors of our models due to the short time frame. It's also advisable to do some error analysis before using our models. The proposed ensemble models and fine-tuned models can be used to analyze the social media content. Although this study includes data from Twitter, the amount of data is not sufficient to analyze the sentiment of Twitter text using our developed system. We used GPUs to run our experiments and heavy use of GPUs contributes to global warming.

\bibliographystyle{plain}
\bibliography{bibfile}

\end{document}

%% file: results.tex

    \centering
    \begin{tabular}{|p{0.08\linewidth}|p{0.1\linewidth}|p{0.3\linewidth}|p{0.08\linewidth}|p{0.08\linewidth}|p{0.08\linewidth}|p{0.08\linewidth}|p{0.06\linewidth}|}
    \hline
        Language & Training Data & Model & Accuracy & Precision & Recall & F1-macro \\ \hline
        \multirow{3}{*}{English} & \multirow{3}{*}{English}& m-BERT (Baseline) & 67.16 & 67.48 & 67.16 & 67.06 \\ \cline{3-7}
         &  & RoBERTa & 70.69 & 71.34 & 70.69 & 70.84 \\ \cline{3-7}
         &  & XLM-RoBERTa & 69.07 & 67.00 & 69.07 & 69.13 \\ \hline
        \multirow{3}{*}{Arabic} & \multirow{3}{*}{Arabic} & m-BERT (Baseline) & 54.21 & 53.76 & 54.21 & 53.08 \\ \cline{3-7}
         &  & AraBERTv02 & 69.79 & 69.96 & 69.79 & \textbf{69.78} \\ \cline{3-7}
         &  & XLM-RoBERTa & 63.89 & 63.63 & 63.89 & 63.74 \\ \hline
        English & English & Majority Voting Ensemble & 70.95 & 71.55 & 70.95 & \textbf{71.03} \\ \hline
        Arabic & Arabic & Majority Voting Ensemble & 66.69 & 66.37 & 66.69 & 66.42 \\ \hline
        English & English & Ensemble model with Feed Forward & 68.91 & 69.26 & 68.91 & 68.59 \\ \hline
        Arabic & Arabic & Ensemble model with Feed Forward & 67.67 & 69.01 & 67.67 & 67.82 \\ \hline
        English & English and Arabic & Ensemble model with multi-head attention Feed Forward & 67.44 & 69.14 & 67.44 & 67.31 \\ \hline
        Arabic & English and Arabic & Ensemble model with multi-head attention Feed Forward & 66.30 & 67.82 & 66.30 & 66.42 \\ \hline
        English & English and Arabic & Ensemble model with Feed Forward & 70.03 & 70.50 & 70.03 & 69.88 \\ \hline
        Arabic & English and Arabic & Ensemble model with Feed Forward & 67.61 & 68.01 & 67.61 & 67.12 \\ \hline
    \end{tabular}